# AI vs. Human - Differentiation Analysis of Scientific Content Generation

Yongqiang Ma [a,*], Jiawei Liu [a,*], Fan Yi [a],

Qikai Cheng [a], Yong Huang [a], Wei Lu [a,**], Xiaozhong Liu [b]

[a] *Wuhan University, China*
[b] *Worcester Polytechnic Institute, USA*

**Abstract:**

Recent neural language models have taken a significant step forward in producing remarkably controllable, fluent, and grammatical text. Although studies have found that AI-generated text is not distinguishable from human-written text for crowd-sourcing workers, there still exist errors in AI-generated text which are even subtler and harder to spot. We primarily focus on the scenario in which scientific AI writing assistant is deeply involved. First, we construct a feature description framework to distinguish between AI-generated text and human-written text from syntax, semantics, and pragmatics based on the human evaluation. Then we utilize the features, i.e., writing style, coherence, consistency, and argument logistics, from the proposed framework to analyze two types of content. Finally, we adopt several publicly available methods to investigate the gap of between AI-generated scientific text and human-written scientific text by AI-generated scientific text detection models. The results suggest that while AI has the potential to generate scientific content that is as accurate as human-written content, there is still a gap in terms of depth and overall quality. The AI-generated scientific content is more likely to contain errors in factual issues. We find that there exists a "writing style" gap between AI-generated scientific text and human-written scientific text. Based on the analysis result, we summarize a series of model-agnostic and distribution-agnostic features for detection tasks in other domains. Findings in this paper contribute to guiding the optimization of AI models to produce high-quality content and addressing related ethical and security concerns.

## 1. Introduction

Recent artificial intelligence (AI) models have taken a significant step forward in generating hyper-realistic content in form such as text, image, and video. However, the ability to create human-like content with unprecedented speed presents additional technical and social challenges. The abuse of AI can cause many issues such as disinformation and information fraud (Zellers et al., 2019). Deepfakes, the realistic videos generated by deep learning, are extremely difficult to distinguish from genuine videos (Lyu, 2020; Mirsky & Lee, 2021). Deepfakes provide information that seems right but is not truthful, which can lead people to acquire false beliefs (Fallis, 2021).

Recent natural language generation (NLG) models show a bright future for AI writing assistants based on large pre-trained NLG models. The coherence, consistency, and grammar of AI-generated text from NLG models have been continuously improved from GPT-2 (Radford et al., 2019), to GPT-3 (Brown et al., 2020), to InstructGPT (Ouyang et al., 2022). The advances in NLG models have empowered writing aids, such as autocomplete, and led to more complex and controllable writing (Sun et al., 2021). AI writing assistant can support people in writing text such as songs, stories, press releases, interviews, essays, and technical manuals (Hutson, 2021). However, the misuse of NLG models raises social concerns in many domains (Crothers et al., 2022). In the education scenario, students could use ChatGPT[1] to cheat in exams or produce essays on a given prompt (Dehouche, 2021; Stokel-Walker, 2022; Susnjak, 2022).

In the scientific domain, there exist challenges in generating scientific text by NLG models. Compared with text in other domains, scientific text should provide novel insights to readers. Moreover, scientists could write diverse

---

scientific text for the same research problem. And the peer review process increases the quality of the manuscript to be published. With the latest NLG models, e.g. ChatGPT, and Galactica (Taylor et al., 2022), there also exists an ethical issue that the highly-fluent human-like generated text could be passed off as the user's works and submitted to conferences or journals. AI conference organizers, including ICML[2], and ACL[3], and journals, such as Nature[4], have updated their authorship policies to address this trend.

As strong as the NLG model is, it still makes mistakes, such as generating literal correct but inconsistent and counterfactual text. Even worse, this content might be used to manipulate public opinion (Liu et al., 2022; Salge et al., 2022). In turn, mass-produced text could contaminate or poison language models (Magar & Schwartz, 2022; Schuster et al., 2021). In the scientific domain, manuscripts generated in this way could pose unprecedented threats and challenges to scientific publishing and research integrity. Identifying AI-generated text can save reviewers' time, ensure the credibility of scientific knowledge, and help people avoid potential misinformation. Moreover, investigating the gap between AI-generated scientific text and human-written scientific text is significant for the research and development, which can help to guide the optimization of AI models and human-AI collaboration in the research process. Therefore, we primarily focus on the scenario in which scientific AI writing assistant is deeply involved and analyze the gap between AI-generated scientific text and human-written scientific text.

Recent works have primarily considered fine-tuning pre-trained models to detect AI-generated text. For instance, with the release of GPT-2, OpenAI also released a detection model[5], which is a RoBERTa-based binary classification model fine-tuned on a dataset consisting of human-written text and GPT-2-generated text. Black et al., (2021) combine source-domain data with in-domain labeled data to solve the problem of detecting GPT-2-generated technical research text. During the Scholarly Document Processing workshop at COLING 2022, Kashnitsky et al., (2022) proposed the task and corresponding dataset on the detection of automatically generated scientific papers, called DagPap22. They adopt GPT-3, GPT-neo, and led-large-book-summary[6] to generate abstracts. Since the prompt templates constructed by DagPap22 do not contain the core topic and scientific structure function information (Dernoncourt & Lee, 2017; Lu et al., 2018), synthetic abstracts collected by DagPap22 are more likely to be problematic and easily detected. More recently, GPTZero[7], mainly based on perplexity, has been introduced to detect ChatGPT-generated text. Previous works mainly rely on fine-tuning end-to-end pre-train models using synthetic data. There is still much room for improvement in the performance and interpretability. Also, they are limited to specific models trained on particular datasets and do not present a realistic or comprehensive scenario where adversary models might be from various domains. As the generation and the detection are a process of a mutual game that presents a spiral and wave-like evolution, the detection also needs to examine the similarity and differences among different language levels, such as syntax, semantics, and pragmatics.

To address the aforementioned issues, we collect a dataset in Computer Science (CS) and Biomedical (Bio) domains. Different from DagPap22 (Kashnitsky et al., 2022) and SynSciPass (Rosati, 2022), the generated abstract is from GPT-3 and ChatGPT, with an optimized prompt containing scientific structure function information, which we will discuss in detail in Section4.2. Then, we conduct a human evaluation on the detection of AI-generated scientific text. Based on the result of human evaluation, we construct a feature description framework to distinguish between AI-generated text and human-written text from syntax, semantics, and pragmatics. Moreover, we also conduct a case study from the view of coherence, consistency, and argument logistics. Similar to other LLMs, we find that the ChatGPT suffers from the hallucination problem, i.e., "reference hallucination", in generated scientific text (Bang et al., 2023). Finally, we employ the feature-based detection method and fine-tuned pre-trained detection method to analyze the gap between AI-generated scientific text and human-written scientific text.

For the feature-based detection method, we utilize the features in writing style, coherence, consistency, and argument logistics on our feature description framework to analyze the similarities and differences between the two types of content. Specifically, the writing style dimension contains token-level features, sentence-level features, perplexity, readability, etc. Then we fine-tuned the GPT-2 output detector model based on our collected dataset. Moreover, we utilize the regression analysis, factor analysis, and model explainability analysis framework to further investigate the gap between AI-generated text and human-written text. Results show that the features in syntax, i.e. writing style, provide the strongest explanation of the feature-based model and the end-to-end fine-tuned pre-train model.

---

[1] https://openai.com/blog/chatgpt/
[2] https://icml.cc/Conferences/2023/llm-policy
[3] https://2023.aclweb.org/blog/ACL-2023-policy
[4] https://www.nature.com/nature/editorial-policies/authorship
[5] https://openai-openai-detector.hf.space
[6] https://huggingface.co/pszemraj/led-large-book-summary
[7] https://etedward-gptzero-main-zqgfwb.streamlit.app/

To summarize, our main contributions are threefold:
- We collect a dataset containing human-written abstracts and AI-generated abstracts, which are generated from LLMs using optimized prompts containing scientific structure function information.
- We investigate feature-based detection method, fine-tune-based detection model, and their corresponding explainability. For the current AI-generated text detection, writing style features from the syntax perspective play a significant role, which shows that there exists a "writing style" gap between the AI-generated scientific text and human-written scientific text. Moreover, we discover that the AI-generated scientific text has a low external inconsistency with the real scientific knowledge world.
- We find that the trained model outperforms the human in distinguishing between the AI-generated scientific text and human-written scientific text, which indicates the desirability of explicitly labeling AI-generated text in the scientific community.

This article is organized as follows: Section 2 demonstrates the application scenarios of NLG models in scientific writing; Section 3 presents a brief literature review; Section 4 elaborates the research methodology; Section 5 describes the result and discussion. The final section concludes this work and suggests directions for future work.

## 2. Application of NLG Models in Scientific Writing

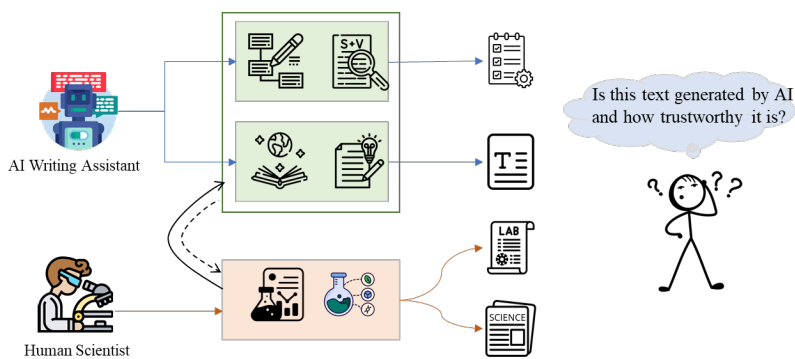

Figure 1 Using the scientist-written content as training data, the AI model can generate human-like and highly fluent scientific text. The reader might ask "Is this paper or abstract generated by AI and how trustworthy is it?" to avoid misinformation when reading a scientific text. As a personal research assistant, the AI writing assistant could also become more and more embedded in the research process.

The NLG model is the core of AI writing assistant, which can improve the efficiency of scientific writing. We enumerate the scenarios in which AI writing assistant is used in scientific writing based on the ACL 2023 policy on AI writing assistant.

**Lightly involved in scientific writing.** The AI writing assistant acts as a language assistant in scientific writing. For example, AI writing assistant can be used for paraphrasing or polishing human-authored content.

**Deeply involved in scientific writing**. The AI writing assistant acts as a partner in scientific writing, collaborating with researchers. For example, the AI writing assistant can be used for describing widely known concepts, generating drafts of related work sections (Li et al., 2022), and even writing new ideas.

For the first scenario, AI writing assistant is acceptable, as it can improve the quality of writing for non-native English speakers. However, for the second scenario, there is a risk of academic fraud and plagiarism when AI writing assistant is deeply involved in scientific writing and even leads scientific writing, such as automatically generating new ideas and text.

As AI writing assistant becomes more and more embedded in the process of scientific writing, people will ask, "Is this paper's text generated by an AI model?" when reading a scientific text as shown in Figure 1. The performance of NLG models is constantly improving. The text generator and the detector are in an adversarial relationship. Training a static detection model based on a static dataset cannot follow the evolution of NLG models and also has difficulty in dealing with domain generalization. Therefore, we investigate the gap between AI-generated text and human-written content to support the long-term solution. Analyzing the mechanism behind AI-generated text in the scientific community can help to optimize the human-AI collaboration in the research process.

## 3. Related Work

In this section, we will first discuss the related work concerning AI generated content, then further introduce and summarize the related works on GPT-generated text detection. Finally, we will introduce the related work on AI-generated content detection in scientific field.

### 3.1. AI Generated Content

In the 1950s when it began, computer-generated content focused on visual art and music(Boden & Edmonds, 2009). Early computer-generated content can be easily distinguished by the general public from human-generated content (Pataranutaporn et al., 2021). With the advancement of artificial intelligence technology, in terms of visual content, the content generated based on techniques such as generative adversarial networks(Goodfellow et al., 2020) and diffusion models (Dhariwal & Nichol, 2021) has become highly realistic. In terms of the misuse of AIGC, e.g. Deepfake (Lyu, 2020), researchers have conducted extensive research from the technical, cognitive, and social perspectives (Fallis, 2021; Guera & Delp, 2018; Karasavva & Noorbhai, 2021; Lee & Shin, 2022; Zhang et al., 2022). The generation of academic texts based on AI models is also an important area of application for AIGC. With regard to the automatic generation of academic papers, Jeremy Stribling developed SCIgen in 2005. SCIgen[8] is a tool for the random generation of computer science research papers based on context-free grammar. SCIgen-generated papers include a complete structure, but their content contains massive errors and nonsense.

Large language models face technical and social challenges while promoting the development of NLP downstream tasks. (Petroni et al., 2019) found that large pre-trained language models can not only learn linguistic knowledge but can also store and simply reason about the world's knowledge from the massive training corpus. To organize scientific knowledge, the MetaAI team unveiled a new large language model called Galactica which can store, combine and reason about scientific knowledge (Taylor et al., 2022). Galactica outperforms existing models on a range of scientific NLP tasks but tends to reproduce prejudice and assert falsehoods as facts. Therefore, MetaAI took down the public demo after three days. The GPT-3 (Brown et al., 2020) model proposed by OpenAI can generate highly fluent text. Given that large language models (e.g. GPT-3) can generate untruthful, toxic, or unhelpful content to the user, Ouyang et al., (2022) use reinforcement learning from human feedback to fine-tune language models for aligning these models with user intent. ChatGPT[9], a sibling model to InstructGPT, has great performance in conversations with humans. It can understand users' instructions better, and generate helpful, trustworthy honest, and harmless text content.

### 3.2. GPT-generated Text Detection

Machine-generated text or AI-generated text is a natural language text generated, rewritten or expanded by a machine or algorithm (Crothers et al., 2022). Clark et al., (2021) found that non-experts could distinguish between GPT3- and human-authored text at random in three domains (stories, news articles, and recipes). A growing number of studies have been conducted to analyze, recognize, and detect AI-generated text, especially GPT-generated text. Current research focuses on two main areas: 1) Human behavior for the recognition of AI-generated text; and 2) Detection model for AI-generated text identification.

#### 3.2.1. Human behavior for the recognition of AI-generated text

---

[8] https://people.sc.fsu.edu/~jburkardt/fun/misc/scigen.html
[9] https://openai.com/blog/chatgpt/

Clark et al., (2021) investigated the non-experts' responses to the AI-generated text (GPT2 and GPT3). Zellers et al., (2019) found that humans rated Grover-generated misinformation as more trustworthy than human-written disinformation. Köbis and Mossink, (2021) conducted an empirical study of people's ability to discern artificial content (GPT-2 produced samples of poems) from human content, and found that participants could not reliably detect GPT2-generated poems. Kreps et al., (2022) carried out experiments to study the influence of AI-generated texts' opinions on foreign policy. Jakesch et al., (2022) found that individuals achieved a low identification accuracy for the AI-generated self-presentations in three social contexts (job applications, online dating, and Airbnb host profiles) and heuristics can improve the identification accuracy. The common conclusion of recent research is that individuals are largely incapable of distinguishing between AI- and human-written text. Additionally, the focus point is different between non-experts and experts when trying to identify the GPT-generated text (Clark et al., 2021), and certain genres (generated recipes) are slightly easier than others (generated stories and news articles) (Dugan et al., 2022).

*3.2.2. Detection model for the AI-generated text identification*

Zellers et al., (2019) proposed the Grover model to generate fake news samples and detect the fake news. After GPT-2 was released, OpenAI proposed the GPT-2-generated text detector that achieved a high F1 score. The detector was fine-tuned based on Roberta in a binary text classification format. Dugan et al., (2020) proposed a boundary-detection task, which identifies article transitions from being human-written to being AI-generated. The boundary-detection task can provide a fine-grained understanding of the hybrid text, which contains human-written content and AI-generated content. Dou et al., (2022) proposed a framework called Scarecrow for machine text detection. Scarecrow has 10 error categories commonly found in an AI-generated news article. An error category prediction model was trained based on Scarecrow, which achieved higher model F1 scores than the human annotators for half of the error span categories. Most of the current research primarily focuses on the text in news text or online text. Mitchell et al., (2023) posed a hypothesis that the LLM-generated text tends to occupy the areas where the log probability function has negative curvature (e.g., local maxima of the log probability). Based on the hypothesis, Mitchell et al., (2023) proposed DetectGPT, which is a zero-shot approach for LLM-generated text detection.

*3.3. AI-generated Scientific Content*

As an important carrier of knowledge in the scientific communication system, the scientific paper is the core of the scientific community, containing the novel insights of researchers. By reading scientific papers, people can obtain credible knowledge about things. AI enables the research process from scientific paper retrieval and scientific paper recommendation to scientific paper writing. AI writing assistants such as ChatGPT are playing an increasing role in the process of scientific writing, which has raised concerns in the scientific community. When AI writing assistants are abused (e.g. using ChatGPT to generate new ideas and new text as a part of the manuscript), there exists a potential for plagiarism and academic fraud. In 2023, AI conferences (ICML, ACL) updated their policy about submitted manuscripts to avoid fake, plagiarized, or fraudulent findings generated by large-scale language models, especially ChatGPT.

Since the release of SCIgen in 2005, researchers have found that fake papers generated by SCIgen were published by academic paper publishers such as Springer and IEEE (Van Noorden, 2014, 2021). To identify machine-synthesized academic papers, Labbé & Labbé, (2013) proposed a text mining tool to detect fake papers. Cabanac and Labbé, (2021) found that the prevalence of SCIgen papers in information and computing sciences is estimated to be 75 per million papers. Additionally, Oberreuter and Velásquez, (2013) quantified the writing style of a text by calculating the word frequency in the text, which was used to detect text fragments in which plagiarism was suspected.

Recently, the text generated based on the pre-trained language model has made great progress in terms of text fluency and text coherence. GPT-3 can even write a paper about itself with the title "Can GPT-3 write an academic paper on itself, with minimal human input?" (Thunström & Steingrimsson, 2022). The NLG models with strong text generation ability can be used to facilitate plagiarism (Dehouche, 2021; Else, 2021). In this challenge, a shared task to detect automatically generated scientific papers (DAGPap22) was proposed in the third workshop on scholarly document processing (Cohan et al., 2022; Kashnitsky et al., 2022). DAGPap22 is formalized as a binary classification task. Rosati, (2022) reframed the binary classification task in DAGPap22 as detecting the type of tool used for generating text because mislabeling a submitted manuscript as AI-generated is harmful to the author(s).

## 4. Methodology

Based on the analysis in Section 2, we primarily focus on the scenario where AI writing assistant is deeply involved in scientific writing. The scientific abstract contains the key findings and summarizes the core information of a scientific paper. As a scenario where AI is deeply involved in academic writing, we use the NLG model to generate the abstract from a given prompt with title information in this work. Specifically, We employ the GPT-3 model proposed by OpenAI as our scientific papers' abstract generator. The source papers of the abstract come from PubMed, ACL, and Arxiv. The information about the abstract is shown in the Table 1.

Table 1 Abstracts information

| Source | Number of Abstracts | Description |
| --- | --- | --- |
| PubMed | 750 | The query term is "covid-19." |
| ACL | 600 | Long papers published in ACL-2022. |
| Arxiv | 1150 | Submitted in Arxiv labeled with cs.AI, cs.CV, cs.CL, and cs.LG. |

### 4.1. Text Generator

To the best of our knowledge, ChatGPT, a sibling model to InstructGPT (Ouyang et al., 2022), is a state-of-the-art natural language generation model. But there is no available API for ChatGPT. Additionally, ChatGPT trades in-context learning performance for dialog history modeling compared with Text-Davinci-003, which is the most powerful GPT-3 model in OpenAI with available API. Therefore, we primarily employ Text-Davinci-003 as our text generator in this work. Moreover, we manually collect small-scale ChatGPT-generated text to compare the difference between GPT3-generated text and ChatGPT-generated text, and for the case study.

### 4.2. Prompt Design

A structured abstract summarizes the key findings reported in a scientific paper. It enables readers to learn about conclusions or how those conclusions were reached without reading the paper in its entirety. We design the prompt for scientific abstract generation based on the scientific structure function of the abstract (Lu et al., 2018). Structured abstracts differ from subject to subject. The scientific papers' abstract generation prompts are shown in Table 2. In our dataset, the generated paper abstract is labelled as fake, and the original abstract of a paper written by humans is labelled as real.

Table 2 Prompt for generating the scientific abstract. The "TITLE" is a placeholder, which is replaced by a paper title when requesting the OpenAI API.

| Domain | Sections in Structured Abstracts | Prompt |
| --- | --- | --- |
| Biology | Background, Objectives, Methods, Results, Conclusions (Dernoncourt & Lee, 2017; Jin & Szolovits, 2018) | Write an abstract for the scientific paper titled with "TITLE" with distinct, labeled sections (Background, Objectives, Methods, Results, Conclusions) |
| Computer Science | Background, Motivation, Methods, Results, Conclusions (Accuosto, 2021; Accuosto & Saggion, 2019) | Write an abstract for the scientific paper titled with "TITLE" with distinct, labeled sections (Background, Motivation, Methods, Results, Conclusions) |

### 4.3. GPT-generated Text Detection

We formalized the GPT-generated text detection as a binary text classification task. Given a text example, the detector classifies the text as entirely human-written or entirely AI-generated. Here, we build the GPT-generated text detection model in the feature-based style and neural network-based style. The scientific paper is the core item of the scientific community. The scientific abstract contains the key findings and summarizes the core information of the scientific paper. Therefore, we primarily focus on the AI-generated scientific abstract text detector.

*4.3.1. Feature-based GPT-generated text detection model*

We construct a feature description framework to distinguish between AI-generated text and human-written text from syntax, semantics, and pragmatics. In our work, the feature is categorized into four dimensions (Writing Style, Coherence, Consistency, and Argument Logistics) in our framework. The feature description framework of this paper is shown in Table 3.

Table 3 Feature description framework

| Perspectives | Dimensions | Features |
| --- | --- | --- |
| Syntax | Writing Style | Token level features (e.g. length of the word, part of speech, function word frequency, and stopword ratio) and sentence level (e.g. length of sentence) |
| Semantics | Coherence | Cosine similarity between abstract sentences |
|  | Consistency | Cosine similarity between title and each sentence in the abstract |
| Pragmatics | Argument Logistics | Self-contradiction, redundant, and commonsense |

- From the perspective of syntax, the GPT-generated text detection task is similar to authorship identification, which classifies a text based on content-independent features called writing style instead of the topic and content (Gamon, 2004, 2004).
- From the perspective of semantics, generating text using an AI model in the scientific field is a controlled text generation task. Inspired by the controlled text generation model evaluation metrics (Ke et al., 2022), we identify the AI-generated text from the viewpoint of coherence and consistency because the AI-generated text is not perfect in terms of coherence and consistency.
- From the perspective of pragmatics, writing a scientific paper is a process of dialogue with the potential reader. Human-written scientific papers are very concerned with the logic of argumentation.

We use authorstyle[10] to extract text writing style features. For the features in argument logistics, we trained an error-type classification model based on the dataset proposed by Dou et al., (2022). We use the SciBERT (Beltagy et al., 2019), a pre-trained model trained on scientific text, to obtain the text embedding to compute the cosine similarity between sentences. For the bartscore, we use the bart-large-cnn model[11] which is fine-tuned on CNN daily mail. The AI-generated text is labeled as 0 and the human-written text is labeled as 1.

*4.3.2. Neural Network-Based GPT-generated Text Detection Model*

The RoBERTa-based OpenAI Detector proposed by OpenAI is trained on the outputs of the 1.5B-parameter GPT-2 model with RoBERTa as the backbone[12]. In our work, we fine-tune the GPT-2 output detector model based on our dataset.

## 5. Result and Discussion

*5.1. Analysis of Human Ability to Distinguish AI-generated Scientific Text*

We conduct an evaluation to study how humans discern whether a scientific text was generated by AI. We gave the evaluator 20 scientific papers abstract and 20 wiki item descriptions, some of which were written by people and some generated by ChatGPT. The AI-generated scientific text is shown in Figure 2. The scientific paper is collected from Arxiv in the computer science category. The wiki items are common concepts in NLP. The evaluators are two

---

[10] https://github.com/mullerpeter/authorstyle
[11] https://huggingface.co/facebook/bart-large-cnn
[12] https://huggingface.co/roberta-base-openai-detector

Ph.D. students with a computer science background. As shown in Figure 3, we asked them to rate the text on a 4-point scale following Clark et al., (2021) and write down the reason for making the choice. We find that the human evaluators achieved a 66% F1 score on the total 40 scientific texts as shown in Table 4. Specifically, the human identification of AI-generated abstracts is more accurate than their identification of AI-generated wiki item descriptions.

1) Definitely human-written
2) Possibly human-written
3) Possibly machine-generated
4) Definitely machine-generated

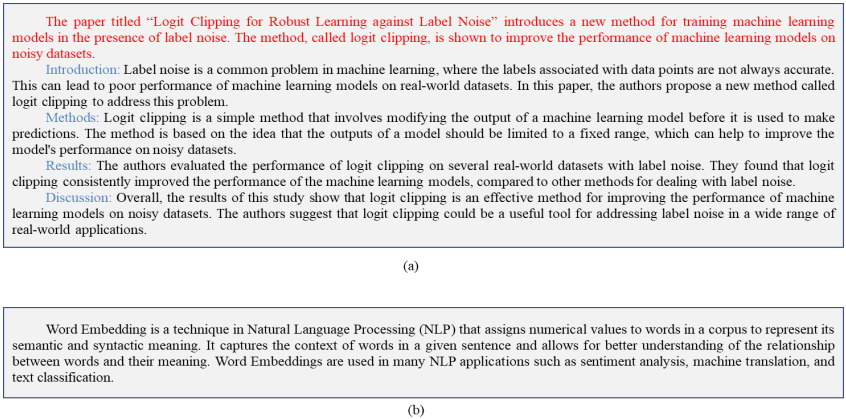

(a)

(b)

Figure 2 (a) is the ChatGPT-generated paper abstract for the paper "Logit Clipping for Robust Learning against Label Noise." For the ChatGPT-generated abstract, we manually removed the undesired part (the blue text and red text) from the ChatGPT-generated text in human evaluation. The text generation prompt is shown in Table 2. (b) is the ChatGPT-generated wiki item description about word embedding. The prompt is "write a short description for "word embedding" in the style of Wikipedia."

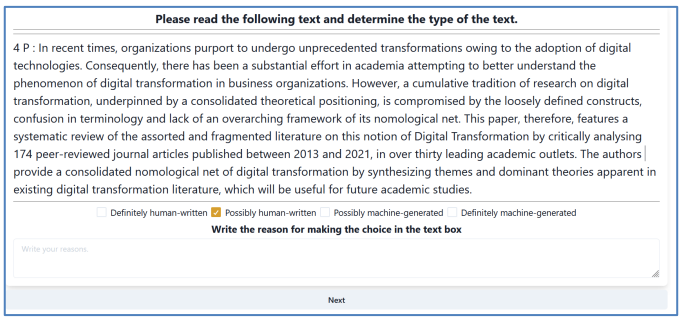

Figure 3 The task interface. The P refers to paper abstract and the W refers to wiki item description.

Table 4 Human performance in identifying AI-generated scientific text

| Text type | Precision | Recall | F1 score |
|---|---|---|---|
| Abstract Text | 76.00% | 75.00% | 74.70% |
| Wiki Item Description Text | 57.50% | 57.50% | 57.50% |
| Total | 66.50% | 66.20% | 66.10% |

We find that evaluators usually focus on the form features such as writing style. There are two reasons for the different performances in identifying the AI-generated abstracts and AI-generated wiki item descriptions.
- The AI-generated abstract is not specific and lacks descriptions of concrete research motivations and methods. Additionally, the AI-generated abstract can not provide a novel insight.
- The wiki text is a part of the dataset, which is used in the pretraining stage of GPT models. Generating the

wiki item description is a process of recalling the text seen during the training stage. Therefore, the generated wiki item text is high quality and similar to the original text, and so humans are unable to identify it as AI-generated.

*5.2. Analysis of Text Perplexity*

The perplexity (PPL) of a language model the is multiplicative inverse of the probability when predicting the following word conditional on the history words. Intuitively, perplexity can be understood as a measure of uncertainty. In our work, we employ SciBERT to compute the text perplexity. As shown in Figure 4(a) and Figure 4(d), the perplexity of ChatGPT-generated text is lower than the perplexity of human-written text. For the ChatGPT-generated text and GPT3-generated text, the difference of perplexity distribution between them is insignificant as shown in Figure 4(b). Moreover, we find that the distributions of the perplexity of GPT3-polished abstracts and Human-generated abstracts are similar as shown in Figure 4 (c)[13].

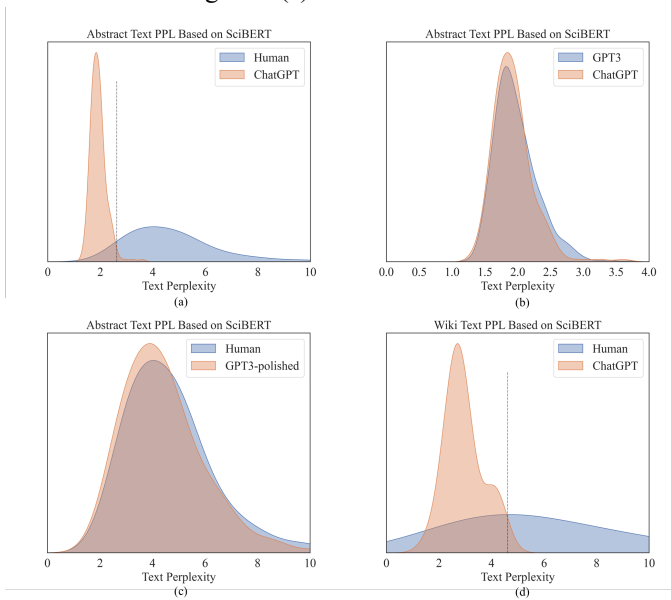

Figure 4 Text perplexity distribution of scientific abstract and wiki item description written by AI models and humans. The vertical line is the position of the threshold for detecting the source type of each scientific abstract. The threshold for scientific abstract is 2.6 as shown in (a). The threshold for wiki item description is 4.6 as shown in (d).

The distribution of perplexity is different between AI-generated and human-written text. Therefore, we set the perplexity threshold as 2.6 for scientific abstract and 4.6 for wiki item text. Specifically, the text is classified as AI-generated when the perplexity is lower than the threshold and as human-written when the perplexity is greater than the threshold. The classification result is shown in Table 5. Using the single text perplexity achieved a 94% F1 score on a scientific abstract and 77% F1 score on wiki item text.

Table 5 The classification result of scientific abstract based on the text perplexity.

|  | Paper Abstract Text | | | | Wiki Item Text | | | |
| --- | --- | --- | --- | --- | --- | --- | --- | --- |
| Type | Precision | Recall | F1 score | Number | Precision | Recall | F1 score | Number |
| AI-generated | 93.3% | 94.9% | 94.1% | 2507 | 71.4% | 100.0% | 83.3% | 25 |

---

[13] The prompt is "polish this paper abstract and keep the original structure and content: ABSTRACT TEXT." The number of GPT3 polished abstracts is 603 in biology and computer science.

| | | | | | | | | |
|---|---|---|---|---|---|---|---|---|
| Human-written | 94.8% | 93.1% | 93.9% | 2491 | 100.0% | 60.0% | 75.2% | 25 |
| Total | 94.0% | 94.0% | 94.0% | 4998 | 85.7% | 80.0% | 79.2% | 50 |

The lower perplexity is caused by the training objective, maximizing the log-likelihood of the token to predict given the current context, in the pre-training stage of GPT models. The training objective forced the model to generate a text with high probability, which resulted in a lower perplexity. Human language capability is used to "do things", such as organizing complex information and persuading someone. Human-written texts are diverse. However, AI-generated texts are required to be as homogeneous as possible with the distribution of the training dataset, thus avoiding the generation of incorrect text.

## 5.3. Case Study

To our best knowledge, ChatGPT is a strong and available text generation model, which has an interface with users. Therefore, we use the ChatGPT-generated scientific text to analysis the coherence, consistency and argument logistics of generated text. Additionally, the case study covers common application scenarios in the research process from the lightly involved in scientific writing to deeply involved, which includes the domain concepts description generation, related work generation and paper review.

### 5.3.1. Coherence and consistency of ai-generated scientific text

As shown in Figure 5a, the generated wiki item about natural language generation is fluent and coherent. But the generated references are wrong. For example, as shown in Figure 5b and Figure 5c, the first and second references are wrong in the ChatGPT-generated text, which has an incorrect author or incomplete title, and the Ref 3 and Ref 4 are completely fake. Additionally, we find that the references in the ChatGPT-generated related work section text are the same as the references in the generated wiki item, as shown in Figure 5 and Figure 6. Moreover, some milestones in natural language generation are missed, such as GPT-series models.

Science needs to be rigorous. The factual knowledge in a scientific text such as journal paper is checked by reviewers. However, the AI-generated scientific text may make up facts compared with human-written scientific text as shown in Figure 6. For example, the generated fake reference is used to demonstrate the related work in natural language generation. Moreover, the generated related work shows the development of natural language generation in a very general perspective. But we can not find the recent progress in natural language generation.

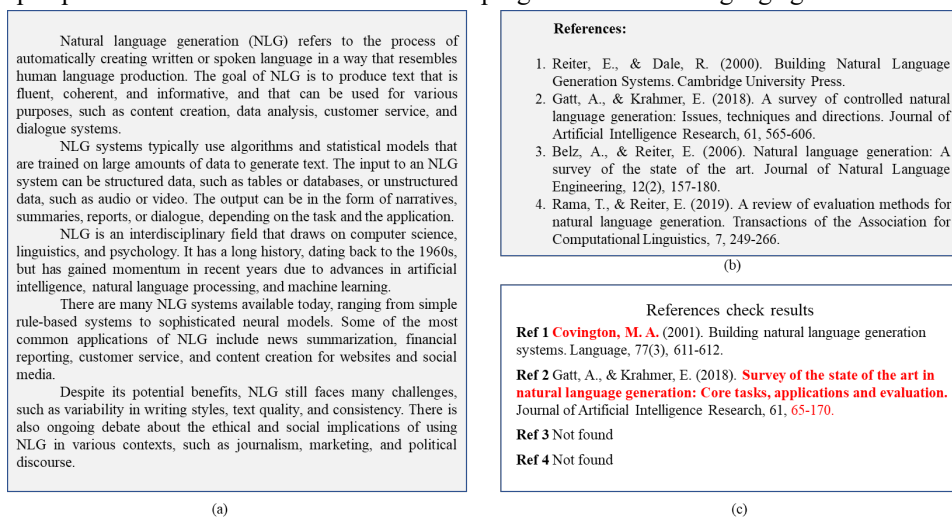

Figure 5 The ChatGPT generated wiki item about natural language generation. (a) is the generated wiki item text; (b) is the wiki item references; (c) is the references check results.

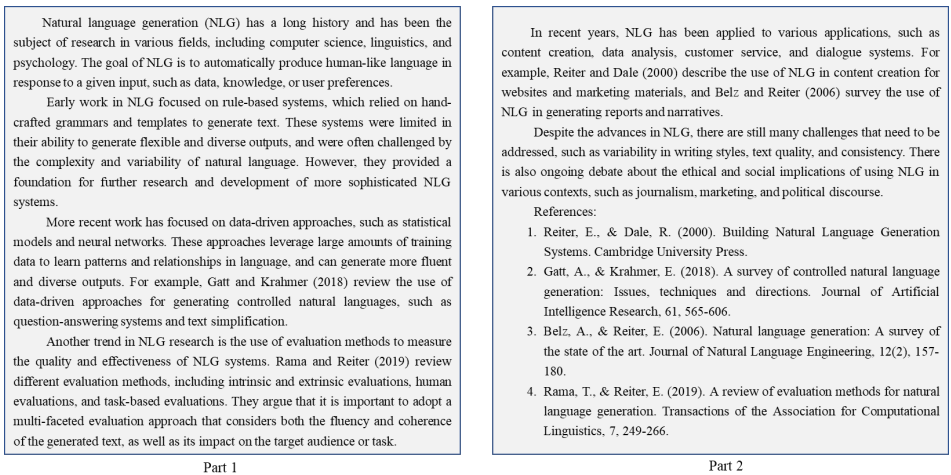

Figure 6 The ChatGPT-generated related work section about natural language generation.

### 5.3.2. Argument logistics of ai-generated scientific text

To show the argument logistics of the ChatGPT model, we require ChatGPT to act as a reviewer for the paper "Med-BERT: pretrained contextualized embeddings on large-scale structured electronic health records for disease prediction" (Rasmy et al., 2021). We find that the ChatGPT-generated review result has good argumentation logic. The pros and cons do not contradict each other. Moreover, the pros and cons can support the conclusion. Finally, ChatGPT make a review decision, which is an "Accept with minor revisions."

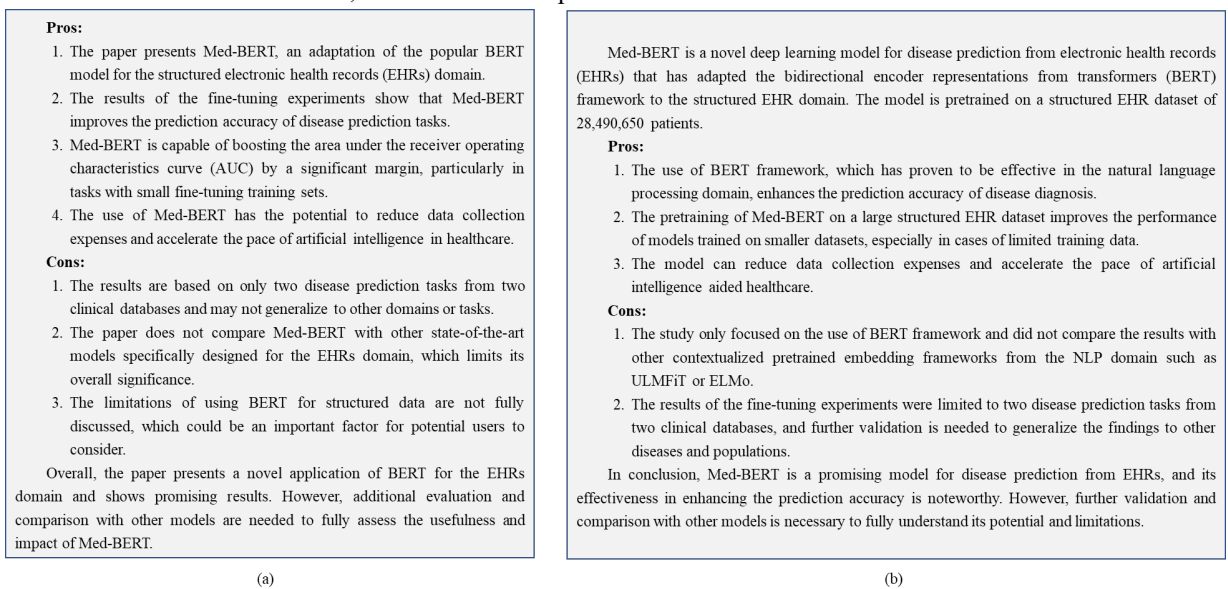

Figure 7 The ChatGPT-generated review result. (a) is the review result based on the paper title and paper abstract. (b) is the review result based on the paper title, abstract, and introduction section.

In conclusion, the AI-generated scientific text has a low external inconsistency with the real scientific knowledge world. The language model is trained on a static training dataset, a snapshot of the dynamic world, which results in the low external inconsistency of the AI-generated scientific text. The number of papers is growing, and retrieval-augmented generation (Lewis et al., 2020) is a valuable solution for the high-quality scientific text generation. Moreover, ChatGPT has the potential to be applied to help researchers improve the quality of their papers.

## 5.4. GPT-generated Scientific Detection Model

### 5.4.1. Logistic regression

To statistically explore the difference between human-written texts and AI-generated texts in syntax, semantics, and pragmatics, we employed a logistic model. The model can provide an interpretable perspective for the difference. We build the logistic model on syntax, semantics, and pragmatics respectively, to show the explanatory power of three perspectives which have four dimensions (writing style, coherence, consistency, and argument logistics). The coefficient and significance level are reported in Table 7. The features with strong correlations are removed to avoid multicollinearity between variables. The VIF value of all features is less than 5, to ensure the effectiveness of the regression.

As shown in Table 7 and Table 6, the features in syntax, i.e. writing style, provide the strongest explanation of the model, which can explain 86.1% of the situation (Pseudo R-square = 0.861). Token features (e.g. average word length, POS tag frequency, punctuation frequency, uppercase frequency), and word features (e.g. function word frequency and average word length) are significant in the task of predicting human-written and AI-generated texts. Text perplexity is significant while average sentence perplexity is not significant.

The third column in Table 7 shows the model built on semantics features achieves a 0.481 Pseudo R-square. Moreover, features in coherence and consistency are significant to predict human-written and AI-generated texts. AI-generated texts have higher coherence with titles but have lower internal consistency. Both the text generated by humans and that generated by AI can deliver semantic information, which results in the logistic regression model based on semantic features having limited explanatory power.

As for the pragmatic level, text redundancy and self-contradiction are significant. Commonsense has been removed because it has multicollinearity with self-contradiction. The text generated by AI has many contradictory parts, but less redundancy, which coincides with lower internal consistency.

Table 6 Information about the logistic regression models

|  | Model 1 | Model 2 | Model 3 | Model 4 |
|---|---|---|---|---|
|  | Only Syntax | Only Semantics | Only Pragmatics | All |
| No. Observations | 3998 | 3998 | 3998 | 3998 |
| Df Residuals | 3978 | 3993 | 3996 | 3971 |
| Pseudo R-squre | 0.861 | 0.4814 | 0.8286 | 0.9378 |
| Log-Likelihood | -385.28 | -1437.1 | -474.99 | -172.24 |
| LL-Null | -2771.1 | -2771.1 | -2771.1 | -2771.1 |
| LLR p-value | 0 | 0 | 0 | 0 |
| F1-score | 0.97 | 0.86 | 0.95 | 0.98 |

Table 7 Coefficient of features in the logistic regression models

| Feature | Model 1 | Model 2 | Model 3 | Model 4 |
|---|---|---|---|---|
|  | Only Syntax | Only Semantics | Only Pragmatics | All |
| Average Word Length | 0.3448 |  |  | -0.151 |
| POS Tag Frequency #ADJ | -0.4511** |  |  | -0.5336* |
| POS Tag Frequency #ADV | -1.1037*** |  |  | -0.9506*** |
| POS Tag Frequency #CONJ | -0.3474*** |  |  | -0.5584*** |
| POS Tag Frequency #NOUN | 0.0527 |  |  | -0.0354 |
| POS Tag Frequency #NUM | -1.1317*** |  |  | -0.7036*** |
| POS Tag Frequency #PRON | -0.6731*** |  |  | -0.5762*** |
| POS Tag Frequency #VERB | 0.4862** |  |  | 0.1967 |
| Flesch Reading Ease | 0.1197 |  |  | 0.3204 |
| Punctuation Frequency#, | -1.0881*** | | | -1.037*** |
| Punctuation Frequency#. | -0.0887 | | | -0.2119 |
| Special Character Frequency#- | -0.3566*** | | | -0.2147 |
| Uppercase Frequency | -0.6413*** | | | -0.5079** |
| Function word Frequency #a | 0.409*** | | | 0.4569** |
| Function word Frequency #in | -0.315*** | | | -0.2711* |
| Function word Frequency #of | 0.0819 | | | 0.0661 |
| Function word Frequency #the | -0.3251** | | | -0.3249 |
| Average Sentences Length | -0.5273*** | | | -0.5916*** |
| Avg Sentences PPL | 0.1069 | | | 0.2253 |
| Text PPL | -6.4385*** | | | -3.5259*** |
| Cos Similarity between Abstract and Title | | -0.6473*** | | -0.4667*** |
| Avg Abstract Sentences Cos Similarity | | 1.254*** | | 0.7407*** |
| Std Abstract Sentences Cos Similarity | | 0.3019*** | | -0.1033 |
| Max Abstract Sentences Cos Similarity | | 1.459*** | | 0.91*** |
| BART Score for Abstract and Title | | 1.1758*** | | 1.1525*** |
| Self-contradiction | | | -5.368*** | -4.2259*** |
| Redundant | | | 1.1935*** | 0.3689* |

Note: * $p < 0.1$, ** $p < 0.05$, *** $p < 0.01$

### 5.4.2. Fine-tuned OpenAI Detector

The RoBERTa large OpenAI Detector is the GPT-2 output detector model trained on the outputs of the 1.5B GPT-2 model[14]. We applied the RoBERTa large OpenAI Detector on our test dataset, which achieved an 88.3 F1 score. Then, we fine-tuned the RoBERTa large OpenAI Detector based on our trained dataset using the transformers[15]. The information about our dataset in shown in Table 8. The learning rate is 4e-7. The batch size is 8. The training epoch is 1. The fine-tuned model achieved a 94.6% F1 score as shown in Table 9.

Table 8 Dataset in the finetuning stage.

| Type | GPT-generated Abstract | Human-written Abstract | Total |
|---|---|---|---|
| Train Dataset | 1502 | 1503 | 3005 |
| Test Dataset | 997 | 997 | 1994 |

Table 9 Result of pre-trained GPT-generated scientific detection models

| | Precision | Recall | F1 Score |
|---|---|---|---|
| OpenAI Detector | 80.7 | 97.4 | 88.3 |
| Our fine-tuned OpenAI Detector | 99.8 | 90 | 94.6 |

LIME is an explanation framework that explains the predictions of any classifier(Ribeiro et al., 2016). LIME explains individual predictions produced by the classifier. We employ the LIME framework to analyze the end-to-end OpenAI detector. As shown in Figure 8, we find that both of the detectors primarily focus on the function word (e.g. the, of, and in). Because the topic is diverse, it is difficult to detect the text type (AI-generated or human-written)

---

[14] https://huggingface.co/roberta-large-openai-detector
[15] https://huggingface.co/docs/transformers/index. The version of transformers is 4.21.1

based on the "content". The detector is primarily focused on the "form" of the text, which is a content-independent feature.

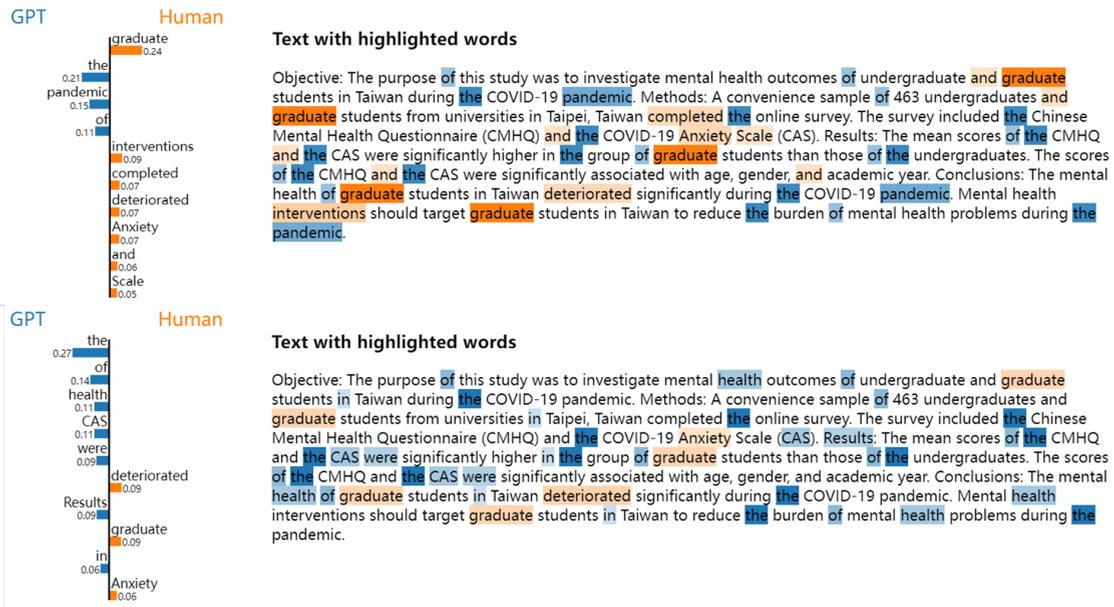

Figure 8 The first figure is the output of LIME based on the RoBERTa large OpenAI Detector. The second figure is the output of LIME based on the fine-tuned RoBERTa large OpenAI Detector. The text in this figure is the GPT-generated abstract for the paper "Investigating mental health outcomes of undergraduates and graduate students in Taiwan during the COVID-19 pandemic."

## 6. Conclusion

The scientific text should provide novel insights to readers compared with text in other domains, such as tweet text and news text. Generating scientific text by NLG models faces challenges and raises many concerns. In this work, we primarily focus on the scenario in which an NLG-based writing assistant is deeply involved in scientific writing. To avoid the abuse of NLG models and the misinformation generated by NLG models in the scientific community, we investigated the gap between scientific AI-generated text and human-written text. Specifically, we first collected the scientific text from the OpenAI API and designed a fine-grained prompt to generate the structured scientific abstract text with the scientific structure functions. Moreover, we conducted a human evaluation to analyze the human ability to distinguish AI-generated scientific text. Then, we constructed a feature description framework to analyze the difference between AI-generated text and human-written text from syntax, semantics, and pragmatics. Based on the constructed framework, we employed the logistic regression model to analyze two types of content. Finally, we fine-tuned the RoBERTa large OpenAI detector based on our dataset and analyze the detection mechanism by an explanation framework.

Generation and detection are a process of the mutual game that presents a spiral and wave-like evolution. The text generator and the detector are in an adversarial relationship. With the development of NLG models, the trained detector based on the static dataset will gradually fail to identify the AI-generated text. Moreover, people will also fail to distinguish between AI-generated scientific text and human-written scientific text. The trained detection models outperform the humans. Based on the proposed feature framework, the trained logistic regression model achieved a high F1 score on AI-generated scientific text detection and is more interpretable than the end-to-end model.

We investigated the gap between AI-generated scientific text and human-written scientific text. We found that 1) the distributions of text generated by humans and AI are significantly different; 2) The AI-generated scientific text, especially the scientific abstract, lacks valuable insights, containing nothing more than generalities; 3) the AI-generated scientific text has a low external inconsistency with the real scientific knowledge world. Our findings can help to optimize scientific text generation models and enhance human-AI collaboration in the research process.

With the development of the NLG model, the difference between AI-generated text and human-written text in terms of syntax will be reduced. The features of semantics and pragmatics will play a significant role in detecting the AI-generated text. Therefore, in the future, we will further study the features of coherence, consistency, and argument logistics in terms of semantics and pragmatics. The number of papers is growing, and the retrieval-augmented generation is a valuable solution for the generation of high-quality scientific text. Moreover, the large language model has the potential to help researchers improve the quality of their research.